\documentclass{bmvc2k}
\usepackage{amssymb}
\usepackage{capt-of}
\usepackage{booktabs}

\title{Unified Representation Learning for Cross Model Compatibility}

\addauthor{Chien-Yi Wang}{chiwa@microsoft.com}{1}
\addauthor{Ya-Liang Chang}{yaliangchang@cmlab.csie.ntu.edu.tw}{2}
\addauthor{Shang-Ta Yang}{shanya@microsoft.com}{1}
\addauthor{Dong Chen}{doch@microsoft.com}{3}
\addauthor{Shang-Hong Lai}{shlai@microsoft.com}{1}

\addinstitution{
 Microsoft Cloud and AI \\
 Taipei, Taiwan
}
\addinstitution{
 National Taiwan University \\
 Taipei, Taiwan
}
\addinstitution{
 Microsoft Research Asia \\
 Beijing, China
}

\runninghead{Wang ET AL.}{URL for Cross Model Compatibility}

\def\etal{\emph{et al}\bmvaOneDot}

\begin{document}
\maketitle
\begin{abstract}
We propose a unified representation learning framework to address the Cross Model Compatibility (CMC) problem in the context of visual search applications. Cross compatibility between different embedding models enables the visual search systems to correctly recognize and retrieve identities without re-encoding user images, which are usually not available due to privacy concerns. While there are existing approaches to address CMC in face identification, they fail to work in a more challenging setting where the distributions of embedding models shift drastically. The proposed solution improves CMC performance by introducing a light-weight Residual Bottleneck Transformation (RBT) module and a new training scheme to optimize the embedding spaces. Extensive experiments demonstrate that our proposed solution outperforms previous approaches by a large margin for various challenging visual search scenarios of face recognition and person re-identification.

\end{abstract}

\section{Introduction}
\label{sec:intro}
\begin{figure}[t]
    \includegraphics[width=\linewidth]{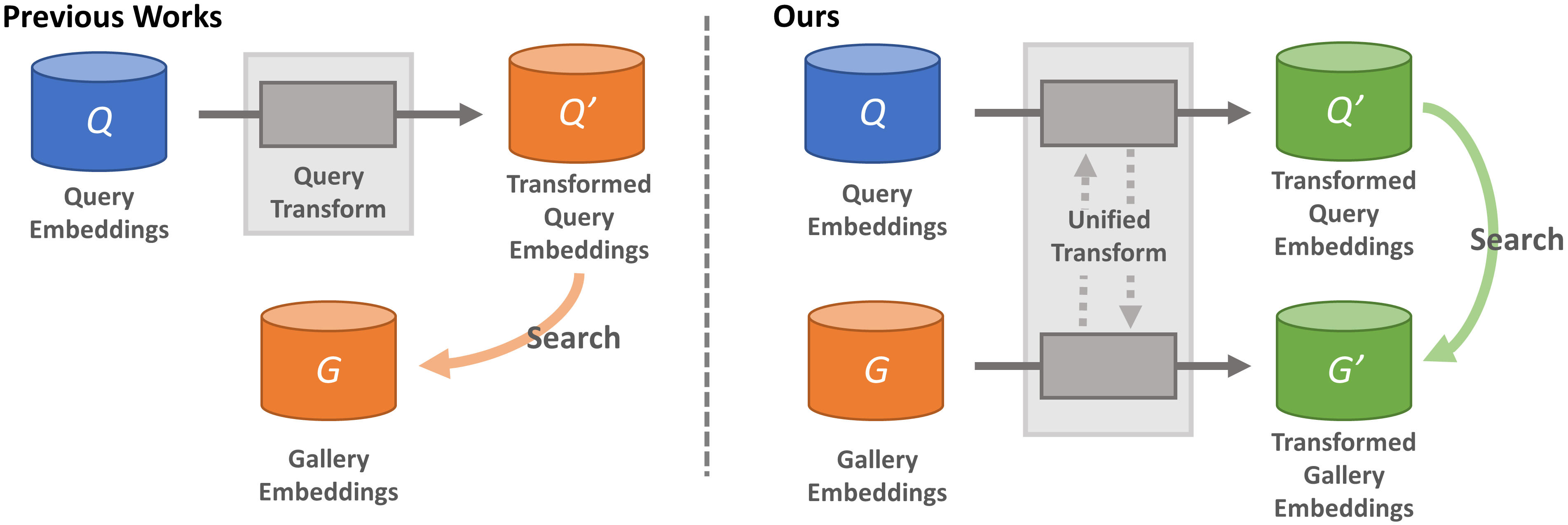}
    \caption{Cross Model Compatibility (CMC) issue takes place while query and gallery embeddings were encoded from different recognition models. Previous approaches address the issue by modeling the transformation between two embedding spaces directly, which has worse performance (Sec.~\ref{sec:exp}) in many scenarios. Our proposed method aims to optimize a new unified embedding space, which achieves better cross-compatibility.}
    \label{fig:figure1}
\end{figure}

Visual recognition and retrieval systems are widely deployed in our lives, such as frictionless physical access~\cite{masi2018deep}, missing person search~\cite{zheng2016person}, and global place recognition~\cite{lowry2015visual}. Most of these systems fall into the open-set visual recognition, which learns models to encode the images into unique embeddings in the high-dimensional vector space. Embeddings of the same class instances are clustered well in the embedding space, and accurate retrieval and recognition are achieved by finding the nearest neighbors in the space. Due to the fast progress of deep neural network architectures~\cite{he2016deep, huang2017densely, howard2017mobilenets, xie2017aggregated}, representation learning techniques~\cite{schroff2015facenet, deng2019arcface, wen2016discriminative} and large labeled training set~\cite{guo2016ms, zheng2015scalable}, diverse embedding models are deployed to meet the requirements for different scenarios. Moreover, embedding models with improved performance are released and updated continuously to achieve a better user experience.

Common visual search applications like person re-identification~\cite{zhou2019omni} and face recognition~\cite{deng2019arcface} have registered \textit{gallery} embeddings from a large number of identities, and test images are encoded from the same embedding model as \textit{query} to perform recognition in the same embedding space. However, under some practical scenarios like upgrading the recognition model or searching across different device models, the system should be able to well address the Cross Model Compatibility (CMC) issue as the embedding spaces from different recognition models are not compatible with each other. 

Re-encoding the gallery images with the same model seems to be a straightforward solution, but the original gallery images may not be stored in the system due to privacy concerns in the industry. Besides re-encoding the images, another feasible approach is to process \textit{gallery} and \textit{query} embeddings directly with another representation learning module to enable the compatibility. Chen \etal ~\cite{chen2019r3} took the first step in this direction to address CMC in face identification, and proposed R$^3$AN to transform the \textit{query} embeddings into the \textit{gallery} embedding space while reconstructing realistic user face images at the middle stage. However, their approach only works when the two face embedding models are similar, but cannot generalize well in other practical scenarios where embedding models differ a lot.

In this work, we study extensively the relationship between embeddings across different embedding models and propose a unified representation learning framework to address the CMC problem, and it shows outstanding performance across various challenging scenarios. Inspired by ResNeXt~\cite{xie2017aggregated}, we propose the light-weight Residual Bottleneck Transformation (RBT) module to learn the embedding transformation very efficiently. Stacking fully connected layers would result in the heavy parameters and gradient vanishing problems during the training. RBT blocks mitigates these issues by skip-connection, channel down-scaling, and path splitting. Instead of transforming the \textit{query} embeddings into the \textit{gallery} embedding space, we propose to transform both \textit{query} and \textit{gallery} embeddings into a unified embedding space. We adopt similarity loss, dual classification loss, and KL loss in the framework to learn a new embedding space which clusters both the transformed \textit{query} and \textit{gallery} embeddings with low intra-subject variation as well as high inter-subject variation. Compared to previous approaches, our paradigm has one more degree of freedom which optimizes the unified embedding space to fit the CMC need better. Moreover, our unified framework generalizes well across various cross model scenarios and open-set visual recognition tasks.

The contributions of this work are summarized as follows: \begin{itemize}
    \item We formalize the Cross Model Compatibility (CMC) problem in the context of visual recognition and retrieval. This new problem aims to model the relationship between embedding spaces from different visual recognition models.  
    \item We propose the light-weight Residual Bottleneck Transformation (RBT) module and a unified learning framework to overcome the CMC problem. The light-weight RBT module mitigates the model convergence issue in previous approaches, and the framework learns the dual transformation for \textit{query} and \textit{gallery} embeddings and achieves better compatibility in the new embedding space. 
    \item The proposed RBT module and the learning framework demonstrate superior performance over previous approaches by up to 9.8\% across challenging scenarios in face recognition and person re-identification tasks.
\end{itemize}

\section{Related Works}
\subsection{Open-set visual recognition}
Face recognition~\cite{masi2018deep, deng2019arcface}, person re-identification~\cite{zheng2015scalable, zhou2019omni}, and image retrieval~\cite{gordo2016deep, tolias2015particular} are popular open-set visual recognition tasks. Deep neural networks (DNNs) are widely applied to learn embedding models that encode each image into an embedding vector. Open-set recognition and retrieval are achieved by computing distances between embedding vectors in the learned embedding space. Some methods train a embedding model by leveraging close-set classification as a surrogate task with various forms of loss functions~\cite{deng2019arcface, Wang2018CosFaceLM, wen2016discriminative}, while others apply metric learning to enforce affinity between embeddings~\cite{schroff2015facenet, Hermans2017InDO}. Those methods provided a basis to train a robust visual embedding model to embed the identity images into the representation vectors. 

\subsection{Learning across domains}
To address distribution change or domain shift issue in computer vision tasks, many domain adaptation~\cite{wang2018deep} techniques are proposed to adapt the output distribution from multiple different modalities. Heterogeneous face recognition (HFR)~\cite{ouyang2016survey} is one of them which aims to match face images acquired from different sources (i.e., different devices, resolutions, or wavelengths) for identification or verification. Because of the domain discrepancy between input image sets, data systhesis~\cite{song2018adversarial,lezama2017not} and domain-invariant feature learning~\cite{kan2016multi,wu2018coupled,he2018wasserstein} techniques are applied to address the problem. Our CMC problem differs in that the discrepancy comes from the embeddings itself instead of input images. While there are many works in domain adaptation addressing the distribution shift between images from different domains, the distribution shift between embeddings from different visual recognition models is not well studied.

\subsection{Compatible representation learning}
Shen \etal~\cite{shen2020towards} propose a backward-compatible representation learning technique to learn visual embedding that is compatible with previous embedding models. However, given a large variety of many different embedding models, it is impossible to train a new embedding model which is backward-compatible with every model. Moreover, the technique cannot apply to address the compatibility between deployed embedding models, which is the usual case in the industry. Chen \etal ~\cite{chen2019r3} raises the Cross Model Face Recognition (CMFR) problem, which is the sub-problem of CMC. The proposed R$^3$AN approach is optimized for face identification and is hard to generalize to other visual search applications. We aim to formulate a more general CMC problem and propose a unified training framework to overcome the limitations in ~\cite{chen2019r3}.


\begin{figure}[t]
    \includegraphics[width=\linewidth]{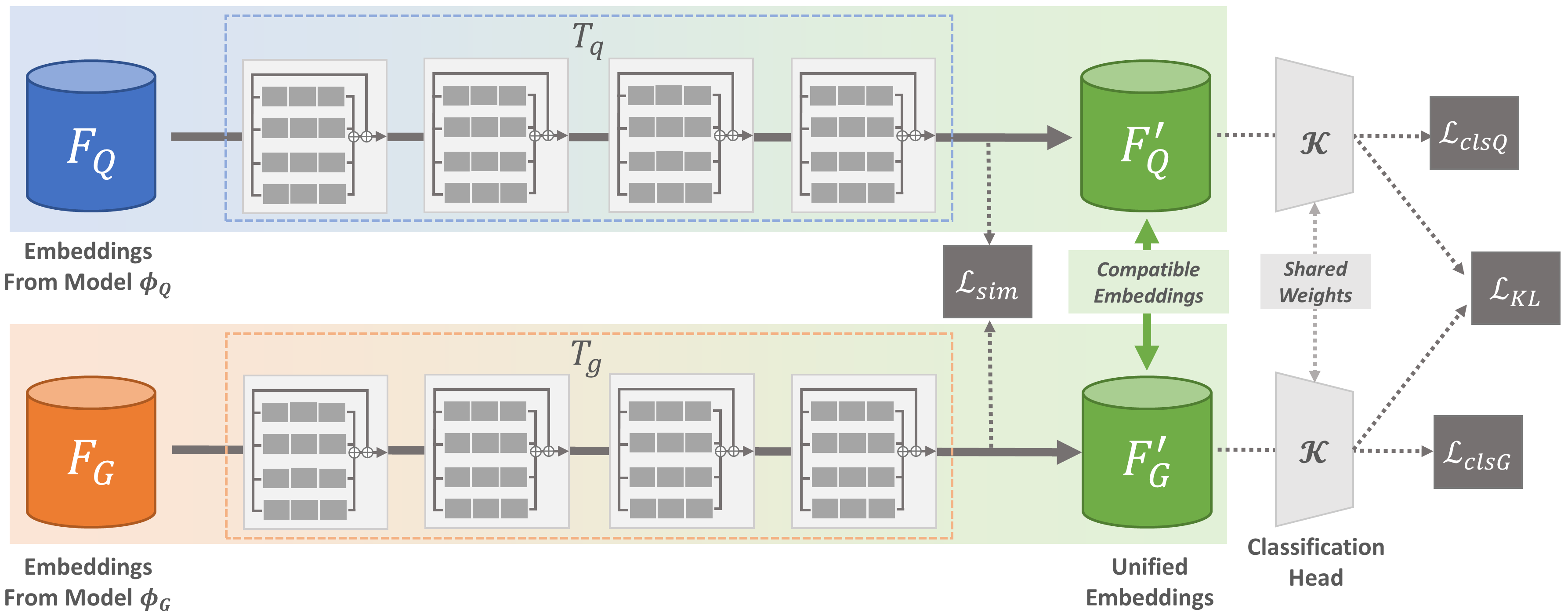}
    \caption{Overview of the proposed unified representation learning framework. Each of the non-linear transformation $T_q$ and $T_g$ between the embedding spaces is composed of four Residual Bottleneck Transformation (RBT) blocks. We optimize the network using three loss functions during the training process: Similarity loss $\mathcal{L}_{sim}$ on the transformed embeddings $F'_Q$ and $F'_G$, dual classification loss $\mathcal{L}_{clsQ}$, $\mathcal{L}_{clsG}$ with the shared classifier, and KL-divergence loss $\mathcal{L}_{KL}$ on the classifier outputs.}
    \label{fig:figure2}
\end{figure}

\section{Proposed Method}
\subsection{Notations and Problem Formulation} \label{notation}
We first define the notations to be used in the CMC problem. The CMC issue takes place while the \textit{query} and \textit{gallery} embeddings are encoded from different embedding models in the context of open-set recognition tasks. We denote the two embedding models as $\phi_Q$ and $\phi_G$. For a group of $N$ sample images, two sets of embeddings are encoded from these images: $F_Q = \{(x_i^q, y_i^q)\}_{i=1}^{N}$ and $F_G = \{(x_i^g, y_i^g)\}_{i=1}^{N}$, where $x_i^q \in \mathbb{R}^{d_q}, x_i^g \in \mathbb{R}^{d_g}$ denotes the embeddings of the $i$-th sample, belonging to the $y_i$-th identity class, and $d_q$ and $d_g$ are the corresponding dimension of embedding models. To address CMC, additional transformation $T_q$ and $T_g$ need to be learned and applied onto $F_Q$ and $F_G$: $F'_Q = T_q(F_Q), F'_G = T_g(F_G)$, and the resulting embedding sets $F'_Q$ and $F'_G$ are compatible with each other. 

\subsection{Residual Bottleneck Transformation (RBT) Module}
\label{sec:rbt}
To address CMC with high dimensional embedding sets, a common choice to model the relationship between embedding spaces is the non-linear mapping by multi-layer perceptron (MLP). However, there are serious drawbacks with MLP: 1) As we show in Sec~\ref{sec:exp}, building a transformation network with deeper MLP cannot lead to better performance. It is highly related to the gradient vanishing issue resulted from a large variation of the weight parameters~\cite{klambauer2017self}. 2) The number of the parameters and FLOPs (floating-points operations) will blow up while stacking high-dimensional transform through fully connected layers.

\begin{figure}
\begin{minipage}[b]{.4\linewidth}
    \includegraphics[width=1.0\linewidth]{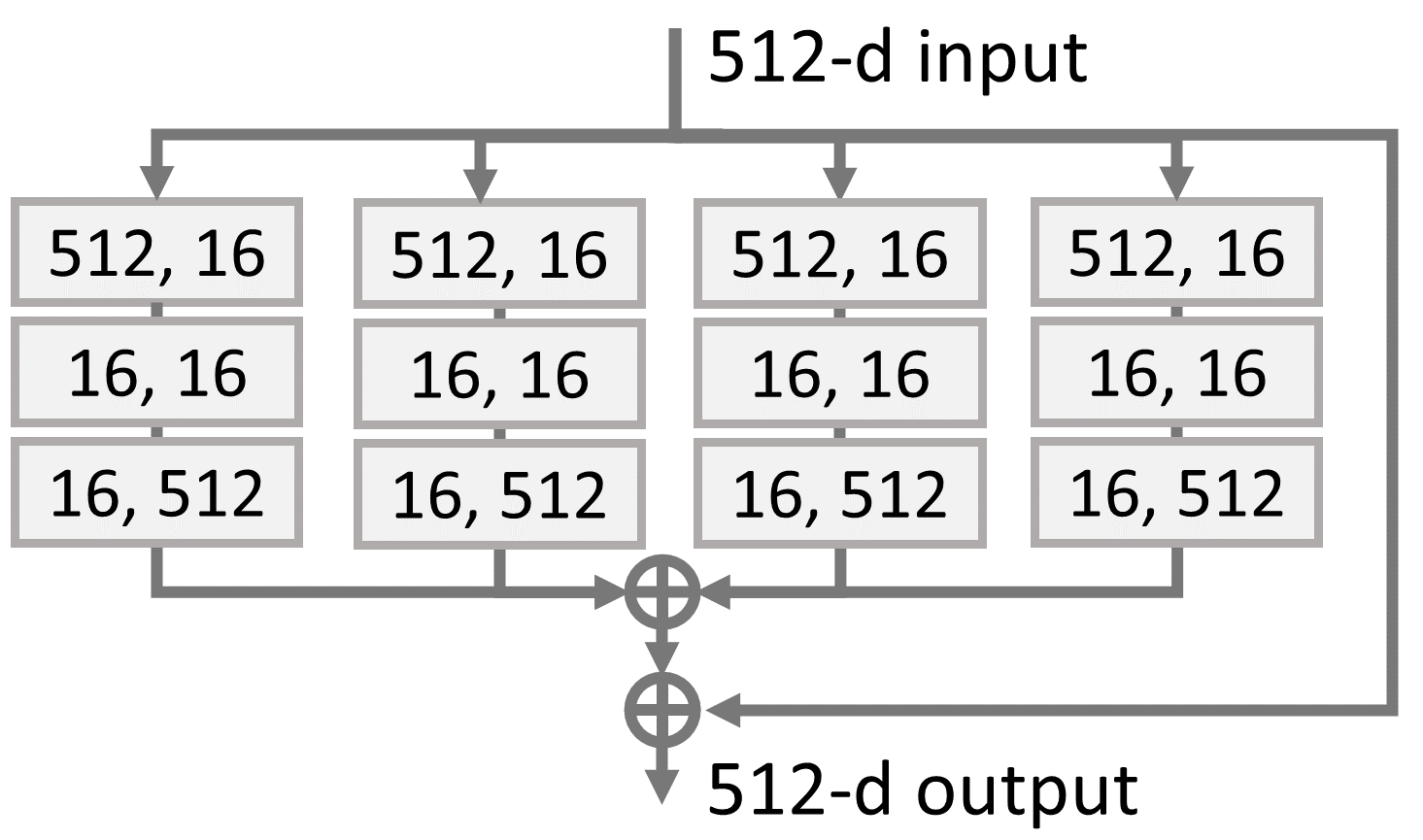}
    \caption{Network architecture of the proposed RBT module. Each small block has \{FC, BatchNorm, ReLU\}, and the number denotes the input and output dimensions.}
    \label{fig:figure3}
\end{minipage}
\hspace{1mm}
\begin{minipage}[b]{.58\linewidth}
    \centering
    \scriptsize
    \begin{tabular}{cccc}
    \toprule
    Methods & $T_q$ & $T_g$ & Losses \\
    \midrule
    MLP baseline & MLP & \textit{Identity} & $\mathcal{L}_{sim}$ \\
    RBT baseline & RBT blocks & \textit{Identity} & $\mathcal{L}_{sim}$ \\
    R$^3$AN~\cite{chen2019r3} & Decoder-Encoder & \textit{Identity} & $\mathcal{L}_{sim}$, $\mathcal{L}_{img}^{*}$ \\
    Ours & RBT blocks & RBT blocks & $\mathcal{L}_{sim}$, $\mathcal{L}_{cls}$, $\mathcal{L}_{KL}$ \\
    \bottomrule
    \end{tabular}
    \vspace*{2mm}
    \captionof{table}{Comparison of different methods for addressing CMC, using the same annotation in the unified framework. \textit{Identity} denotes the identity transformation. $\mathcal{L}_{img}^{*}$ consists of the reconstruction and adversarial losses in~\cite{chen2019r3} which require face input images for training.}
    \label{tab:compare0}
\end{minipage}
\end{figure}

To overcome the above challenges, we propose the Residual Bottleneck Transformation (RBT) module. The detailed architecture of the module is shown in Fig~\ref{fig:figure3}. The RBT module exploits the split-transform-merge strategy, which is widely used in the backbone of Convolutional Neural Networks (CNNs)~\cite{xie2017aggregated, szegedy2015going}, to save the parameters and operations while keeping the representation power. The input is split into four paths of low-dimensional embeddings (bottleneck), transformed by another module, and merged by concatenation. It also mitigates the gradient vanishing issue by the residual skip-connection~\cite{he2016deep}. We leverage the RBT module to build up our strong baseline and the unified learning framework.

\subsection{Unified Representation Learning Framework}
We propose a unified representation learning framework (Fig.~\ref{fig:figure2}) to learn better transformation $T_q$ and $T_g$ to address CMC. The motivation is to learn an unified embedding space which leverage information from both~\textit{query} and \textit{gallery} embedding spaces. To optimize the compatibility between the transformed embedding sets $F'_Q$ and $F'_G$, the unified embedding space needs to be discriminative enough to separate the identities. Firstly, we employ several RBT module blocks into $T_q$ and $T_g$ to encourage efficient non-linear embedding transformation. Secondly, the transformed embedding sets $F'_Q$ and $F'_G$ are passed into the shared classification head $h$ to classify the identities. During the training stage, the closed-set identity classification is treated as a surrogate task that provides a supervision signal to improve generalization ability of the transformed embeddings. The classification head $h$ is shared to ensure the compatibility between the two embedding spaces. Several loss functions are applied to train the network:

\paragraph{\textbf{Similarity Loss.}} Given the sets of the transformed embedding $F'_Q = T_q(\{(x_i^q, y_i^q)\}_{i=1}^{N})$ and $F'_G = T_g(\{(x_i^g, y_i^g)\}_{i=1}^{N})$, the embeddings from the same input sample are enforced to be closed in the unified embedding space as they contain the similar semantic information. We apply $\mathcal{L}_{2}$ losses as the similarity loss on the transformed embeddings:
\begin{equation}
\begin{aligned}
 \mathcal{L}_{sim} = \sum_{i=1}^N | T_q(x^q_i) - T_g(x^g_i)|_2
\end{aligned}
\label{eq:sim}
\end{equation}

\paragraph{\textbf{Dual Classification Loss.}} The classification head $\mathcal{K}$ classify the identities of the transformed embeddings to provide the supervision signal. By sharing the classification head, it enforces the embedding spaces of $F'_Q$ and $F'_G$ to be aligned with each other. The choice of the head $\mathcal{K}$ (e.g. Softmax, Cosine~\cite{Wang2018CosFaceLM}, AM-Softmax~\cite{wang2018additive}, Arcface~\cite{deng2019arcface}) depends on the visual recognition task, and we denote the weights and biases in the fully connected head as $\textbf{W}$ and $\textbf{b}$. The dual classification loss is calculated as follows:
\begin{equation}
\begin{aligned}
 \mathcal{L}_{cls} = \mathcal{L}_{clsQ} + \mathcal{L}_{clsG} = - \frac{1}{N} \sum_{i=1}^N (log\mathcal{K}(T_q(x^q_i), y_i^q, \textbf{W}, \textbf{b}) + log\mathcal{K}(T_g(x^g_i), y_i^g, \textbf{W}, \textbf{b}))
\end{aligned}
\label{eq:cls}
\end{equation}

\paragraph{\textbf{KL-divergence Loss.}} We also penalize the KL-divergence of the classifier output probabilities between transformed \textit{query} and \textit{gallery} embedding from the same input sample, and the KL-divergence loss is calculated as follows:
\begin{equation}
\begin{aligned}
 \mathcal{L}_{KL} = \sum_{i=1}^N KL((\mathcal{K}(T_q(x^q_i), y_i, \textbf{W}, \textbf{b}), \mathcal{K}(T_g(x^g_i), y_i, \textbf{W}, \textbf{b}))
\end{aligned}
\label{eq:kl}
\end{equation}

The total loss for the proposed unified training framework is:
\begin{equation}
\begin{aligned}
\mathcal{L}_{total} = \lambda_{1} \mathcal{L}_{sim} + \lambda_{2} \mathcal{L}_{cls} + \lambda_{3} \mathcal{L}_{KL}
\end{aligned}
\label{eq:loss_total}
\end{equation}
where $\lambda_{1}$, $\lambda_{2}$, and $\lambda_{3}$ are weights for the $\mathcal{L}_{sim}$, $\mathcal{L}_{cls}$, and $\mathcal{L}_{KL}$, respectively. In all the experiments, we empirically set $\lambda_{1} = 1.0$, $\lambda_{2} = 1.0$, $\lambda_{3} = 0.25$.

\section{Experiments}
To verify the effectiveness of our proposed unified learning framework for cross model compatibility, we design a series of experiments with different cross model scenarios. We assess our proposed RBT module and the unified representation learning framework for the face identification problem and compare the compatibility performance with other approaches. We prepare and train various commonly used face embedding models to provide challenging cross model recognition tasks. Then we extract embeddings of the samples in the training dataset from prior face embedding models to address CMC between these embedding sets.

\subsection{Prior Face Embedding Models}
To have a holistic comparison of CMC and verify the generalization ability of our proposed framework, we prepare various prior embedding models with different network backbones and different training loss for identity classification. In real world applications, those in-play embedding models were not all using the same training loss, which would lead to very different embedding distributions, and the modeling of the relationship between embedding spaces could be more challenging. In total there are six prior face embedding models (Table~\ref{tab:prior}), which were all trained on the same training dataset MS1M-retinaface~\cite{deng2019lightweight}, and were evaluated on the MegaFace~\cite{kemelmacher2016megaface} (challenge 1 with FaceScrub as probe set). The MS1M-retinaface dataset is based on MS1M dataset~\cite{guo2016ms} and refined by RetinaFace~\cite{deng2019retinaface} face detector. The output dimension is 512 for all face embedding models.

\begin{table}[t]
\begin{minipage}[t]{.55\linewidth}
    \centering
    \scriptsize
    \caption{Details of prior face embedding models. Network: backbone architecture of the model. Name: abbreviation for the model. Loss: training loss for identity classification. FLOPS: floating-point operations, in \# of multiply-adds. Top-1 Acc: top-1 identification accuracy on MegaFace~\cite{kemelmacher2016megaface} with 1M distractors.}
    \label{tab:prior}
    \vspace{7mm}
    \begin{tabular}{ccccc}
        \toprule
        Network & Name & Loss & FLOPs & Top-1\\
        \midrule
        ResNet-100~\cite{he2016deep} & R100 & ArcFace & 24G & 98.90 \\ 
        MobileFaceNet~\cite{chen2018mobilefacenets} & Mb & ArcFace & 933M & 95.50 \\
        DenseNet-290~\cite{huang2017densely} & Dns & AM-Softmax & 25G & 98.63 \\
        ProxylessNAS~\cite{cai2018proxylessnas} & Pxy & AM-Softmax & 639M & 96.04 \\
        ResNet-100~\cite{he2016deep} & R100s & Softmax & 24G & 87.83 \\
        ResNet-50~\cite{he2016deep} & R50s & Softmax & 12.6G & 89.84 \\
        \bottomrule
    \end{tabular}
\end{minipage}   
\hspace{3mm}
\begin{minipage}[t]{.4\linewidth}
    \scriptsize
    \caption{Comparison between different model architectures in $R100 \rightarrow Mb$ scenario. MLP($n$) represents the MLP model with $n$ hidden layers. RBT($n$) represents the model with $n$ RBT blocks.}
    \vspace{1mm}
    \label{tab:different-arch}
    \centering
    \begin{tabular}{cccc}
        \toprule
        Models & params & FLOPs & Top-1\\
        \midrule
        MLP(1) & 0.53M & 0.52M & 96.19 \\
        MLP(2) & 0.79M & 0.79M & \textbf{96.30}\\
        MLP(3) & 1.05M & 1.05M & 95.13 \\
        MLP(4) & 1.31M & 1.31M & 71.27 \\
        \midrule
        RBT(1) & 0.33M & 0.33M & 96.70 \\
        RBT(2) & 0.40M & 0.40M & 96.74 \\
        RBT(3) & 0.47M & 0.47M & 96.78 \\
        RBT(4) & 0.54M & 0.54M & \textbf{96.93} \\
        \midrule
        R$^3$AN~\cite{chen2019r3} & 6.64M & 194.48M & 97.04 \\
        Ours & 1.08M & 1.08M & \textbf{97.58} \\
        \bottomrule
    \end{tabular}
\end{minipage}  
\end{table}

\subsection{Baseline approaches} 
\label{sec:mlp}
The naive baseline approach is to learn the transformation $T_q$ between \textit{query} and \textit{gallery} embedding models with multi-layer perceptron (MLP) using the $\mathcal{L}_{2}$ similarity loss. We denoted this approach as the \textbf{MLP baseline}. As discussed in Sec.~\ref{sec:rbt}, we proposed the Residual Bottleneck Transformation (RBT) module to overcome the limitation while learning the high-dimensional embedding mapping with MLP. Therefore, we also build a strong \textbf{RBT baseline}, which replace the MLP with RBT blocks, while using the same $\mathcal{L}_{2}$ similarity loss. All the approaches for comparison are summarized under the same framework in Table~\ref{tab:compare0}.   

\subsection{Implementation details}
\label{sec:implement}
We implement the baselines and proposed training framework using Pytorch. The MLP baseline in all experiments is built by two hidden layers, which reaches the best performance, with 512-dimension output in each hidden layer. In the RBT baseline and the proposed solution, we employ four blocks of RBT in $T_q$ and $T_g$. We choose Arcface~\cite{deng2019arcface} as the classification head $\mathcal{K}$ in the face identification experiments with the margin parameters $s = 64$ and $m = 0.5$. The baseline models and the proposed framework are all trained with learning rate starting from 0.1 and divided by 10 after 20 and 25 epochs, and terminate the training after 30 epochs. We also re-implement and train the previous work R$^3$AN~\cite{chen2019r3} with the same protocol described in the paper.

\subsection{Experimental results}
\label{sec:exp}
\paragraph{Evaluation protocol.} In the following experimental results, we use the face identification task MegaFace~\cite{kemelmacher2016megaface} challenge 1 with facescrub as the probe and 1M distractors to evaluate the compatibility between prior embedding models. The experiment is denoted as $M_q \rightarrow M_g$ if we use transformed embeddings from $M_q$ as probe and transformed embeddings from $M_g$ as distractors in the evaluation. The top-1 face identification accuracy is reported in the table.

\paragraph{Effects on different architectures.} In Table~\ref{tab:different-arch}, we demonstrate the difference in model parameters, FLOPs and CMC performance under the $R100 \rightarrow Mb$ scenario. We observe that we cannot build a deep MLP model as the CMC performance drop significantly with increasing hidden layers. It demonstrates that the MLP model with more parameters suffers from gradient vanishing issue. Note that the MLP baseline reported in R$^3$AN~\cite{chen2019r3} is much lower than expected as it built the MLP with the same parameters as the R$^3$AN model. The proposed RBT module mitigates the issues of MLP, and performs better as we increase the number of blocks. In the following experiments, we use MLP with two hidden layers and RBT with four blocks as our strong baselines. Our proposed RBT baseline and unified framework both exhibit comparable results with R$^3$AN~\cite{chen2019r3}, but with only \textbf{16.27\%} and \textbf{0.56\%} of the parameters and FLOPs, respectively.

\begin{table}[t]
    \scriptsize
    \caption{CMC results on embedding models with similar distribution. ($M_q$, $M_g$) denotes the original identification accuracy of the (\textit{query}, \textit{gallery}) embedding models.}
    \vspace{1mm}
    \label{tab:compare1}
    \centering
    \begin{tabular}{ccccccc}
        \toprule
        & R100$\rightarrow$Mb & Dns$\rightarrow$Pxy & R100s$\rightarrow$R50s & Mb$\rightarrow$R100 & Pxy$\rightarrow$Dns & R50s$\rightarrow$R100s \\
        Methods & (98.90, 95.50) & (98.63, 96.04) & (89.84, 87.83) & (95.50, 98.90) & (96.04, 98.63) & (87.83, 89.84) \\
        \midrule            
        MLP baseline & 96.30 & 91.47 & 87.42 & 92.98 & 93.75 & 87.30 \\
        RBT baseline & 96.93 & 94.57 & 87.44 & 96.95 & 95.00 & 87.94 \\
        R$^3$AN~\cite{chen2019r3} & 97.04 & 94.97 & 86.55 & 96.75 & 96.10 & 86.21 \\
        Ours                   & \textbf{97.58} & \textbf{97.27} & \textbf{91.23} & \textbf{97.26} & \textbf{96.83} & \textbf{91.01} \\
        \bottomrule
    \end{tabular}
\end{table}

\begin{table}[t]
    \scriptsize
    \centering
    \caption{CMC results on embedding models with large distribution shift. ($M_q$, $M_g$) denotes the original identification accuracy of the (\textit{query}, \textit{gallery}) embedding models.}
    \vspace{1mm}
    \label{tab:compare2}
    \begin{tabular}{cccccccc}
        \toprule
        & R100$\rightarrow$R50s & R50s$\rightarrow$Dns & Dns$\rightarrow$Mb & Mb$\rightarrow$R100s & R100s$\rightarrow$Pxy & Pxy$\rightarrow$R100 \\
        Methods & (98.90, 87.83) & (87.83, 98.63) & (98.63, 95.50) & (95.50, 89.84) & (89.84, 96.04) & (96.04, 98.90) \\
        \midrule            
        MLP baseline & 85.84 & 58.26 & 94.57 & 86.21 & 75.41 & 84.12 \\
        RBT baseline & 87.85 & 87.86 & 96.56 & 86.26 & 87.07 & 96.43 \\
        R$^3$AN~\cite{chen2019r3} & 88.90 & 87.90 & 96.68 & 86.87 & 86.04 & 82.34 \\
        Ours & \textbf{95.09} & \textbf{92.67} & \textbf{97.14} & \textbf{92.40} & \textbf{93.07} & \textbf{96.59} \\
        \midrule
          & R50s$\rightarrow$R100 & Dns$\rightarrow$R50s & Mb$\rightarrow$Dns & R100s$\rightarrow$Mb & Pxy$\rightarrow$R100s & R100$\rightarrow$Pxy\\
        Methods & (87.83, 98.90) & (98.63, 87.83) & (95.50, 98.63) & (89.84, 95.50) & (96.04, 89.84) & (98.90, 96.04)\\
        \midrule            
        MLP baseline & 89.76 & 81.68 & 83.41 & 85.48 & 77.27 & 89.28\\
        RBT baseline & 90.15 & 86.84 & 93.54 & 87.86 & 84.46 & 96.45\\
        R$^3$AN~\cite{chen2019r3} & 87.44 & 85.00 & 85.36 & 87.81 & 82.72 & 86.13\\
        Ours & \textbf{94.58} & \textbf{93.56} & \textbf{96.65} & \textbf{92.75} & \textbf{92.80} & \textbf{97.19}\\
        \bottomrule
    \end{tabular}
    \vspace{-1em}
\end{table}

\paragraph{Comparisons of different CMC approaches.}
Table~\ref{tab:compare1} shows comparisons of CMC results, under the scenarios where two prior embedding models have similar distributions, as they were trained with the same classification loss. By optimizing the unified embedding space for two embedding sets, our proposed framework exhibit greater ability to address CMC than other approaches. Baseline and R$^3$AN~\cite{chen2019r3} approaches, which only transform one side of the embeddings, can achieve good compatibility results as the distribution changes between embedding models are fairly mild. Notably, in every case, the RBT baseline achieves comparable results with R$^3$AN~\cite{chen2019r3}, which suggests that we do not need complex network architecture to model the distribution shift between embedding sets.  

We further evaluate the CMC approaches with scenarios where two prior embedding models were trained with different classification losses. The comparison results are shown in Table~\ref{tab:compare2}. We observe that the training of R$^3$AN~\cite{chen2019r3} is more unstable than the previous experiments, and requires more tweaking to achieve comparable results. MLP baseline cannot achieve comparable results under these scenarios, as it suffers from the network capacity and training instability. Under such challenging scenarios, our solution performs significantly better than RBT baseline and R$^3$AN~\cite{chen2019r3} by a large margin, which suggests that the general CMC is better addressed by optimizing another embedding space to adapt embeddings from different distributions. In the scenarios of (Dns, Mb) and (Pxy, R100), the RBT baseline can still produce comparable results, which indicates that embedding distributions supported by AM-softmax~\cite{wang2018additive} and Arcface~\cite{deng2019arcface} do not differ a lot.

\paragraph{Effects on the training losses.} We conduct ablation studies on the proposed approach with different training loss combinations and summarize the results in Table~\ref{tab:ablation}. The classification losses by the shared classifier contribute the most for optimizing the unified embedding space, as it encourages the transformed embeddings to be more discriminative on the identities. The similarity loss and the KL loss also improve the identification accuracy by 0.32\% and 0.25\%. Lastly, the proposed method reaches its summit by employing all the losses. 

\paragraph{Transformation cost.} One may raise concerns about the proposed approach as it requires additional transformation on the \textit{gallery} embedding sets. In practical applications, we only need \textit{one-time} transformation on the existed gallery embedding sets. By leveraging the light-weight RBT modules with only 0.54M FLOPs, the transformation on 1M gallery embeddings only takes 0.06 second on a single TITAN RTX GPU, which is very efficient. Moreover, for each on-the-fly \textit{query} transformation, it saves more than 99\% of floating points operations compared with R$^3$AN~\cite{chen2019r3}.        


\subsection{Experiments on Person Re-identification} 
To verify the effectiveness of the unified framework, we validate the proposed method on the person re-identification task using the Market-1501~\cite{zheng2015scalable} benchmark. There are 751 and 750 identities in the training and testing dataset, respectively. We choose three pretrained person embedding models~\cite{zhou2019torchreid} in our CMC experiments: OSNet-1.0 (OS100), OSNet-0.25 (OS25), and MobilenetV2 (Mb). These models differ in the network backbone and the embedding dimensions, that the OSNet has 512-dim, and MobilenetV2 has 1280-dim. In our experiments, to accommodate different input dimensions, the down-scaling dimension in each path of the RBT module is set to $\frac{d_{in}}{32}$. During the training stage of the unified framework, the classification head $\mathcal{K}$ employs Softmax with label smoothing~\cite{zhou2019omni}, which is commonly used in the training of person re-identification models. Search mean average precision (mean AP) is used as the evaluation metric. 

Table~\ref{tab:reid} demonstrates that our proposed RBT module and the unified learning framework can address the CMC issue in the person re-identification task. Besides, our method performs significantly better than MLP and RBT baselines for all scenario, which suggests that the unified learning scheme can better optimize the compatibility between different embedding models. In the future works, larger person re-identification datasets can be leveraged for further improvements on CMC.


\section{Conclusions}
We have presented a unified learning framework for addressing cross model compatibility (CMC) problem in the context of visual search and recognition applications. Our framework robustly optimizes a unified embedding space that adapts embedding distributions from two different embedding models to address CMC. Besides, we proposed a light-weight RBT embedding transformation module to facilitate the training stability and inference efficiency. Based on experimental results, we show that the proposed module and unified framework performs significantly better than previous approaches by a large margin under challenging scenarios in face identification and person re-identification.     

\begin{table}[t]
\begin{minipage}[t]{.35\linewidth}
    \centering
    \scriptsize
    \caption{Ablation experiments on the training losses in the R100$\rightarrow$Mb scenario.}
    \label{tab:ablation}
    \vspace{3mm}
    \begin{tabular}{cccc}
        \toprule
        $\mathcal{L}_{cls}$ & $\mathcal{L}_{sim}$ & $\mathcal{L}_{KL}$ & Top-1\\
        \midrule
        \checkmark & & & 97.10 \\
        \checkmark & \checkmark & & 97.42 \\
        \checkmark & & \checkmark & 97.35 \\
        \checkmark & \checkmark &  \checkmark & \textbf{97.58} \\
        \bottomrule
    \end{tabular}
\end{minipage}
\hspace{1mm}
\begin{minipage}[t]{.6\linewidth}
    \scriptsize
    \centering
    \caption{CMC results on person re-identification. Search mean average precision (mAP) is reported.}
    \label{tab:reid}
    \vspace{1mm}
    \begin{tabular}{cccc}
        \toprule
        & OS100 $\rightarrow$ OS25 &  OS100 $\rightarrow$ Mb & OS25 $\rightarrow$Mb \\
        Methods & (82.6, 75.0) & (82.6, 67.3) & (75.0, 67.3)  \\
        \midrule            
        MLP & 67.1 & 46.7 & 38.4 \\
        RBT & 68.0 & 58.9 & 54.1  \\
        Ours & \textbf{74.1} & \textbf{67.3} & \textbf{62.9} \\
        \midrule
        & OS25 $\rightarrow$ OS100 & Mb $\rightarrow$ OS100 & Mb $\rightarrow$ OS25 \\
        Methods & (75.0, 82.6) & (67.3, 82.6) & (67.3, 75.0) \\
        \midrule            
        MLP & 66.1 & 57.1 & 51.3 \\
        RBT & 67.9 & 59.4 & 54.9 \\
        Ours & \textbf{76.3} & \textbf{65.3} & \textbf{59.1} \\
        \bottomrule
    \end{tabular}
\end{minipage}  
\end{table}

\bibliography{egbib}
\end{document}